\documentclass[9pt]{article} 

\usepackage{times}

\usepackage{epsfig}
\usepackage{psfrag}
\usepackage[latin1]{inputenc}
\usepackage{amsmath,amssymb} 
\usepackage[sort,compress,space]{cite}
\usepackage{url}
\usepackage[algoruled,vlined]{algorithm2e}

\newcommand{\R}[1]{\ensuremath{\mathbb{R}^{#1}}}
\newcommand{\tr}{^{\!\top}}
\newcommand{\inv}{^{-1}}

\begin{document}

\title{Sampling Strategies for Path Planning \\ under Kinematic Constraints}

\author{Josep M. Porta$^\dag$ and L\'eonard Jaillet$^\ddag$ \\
\\
$^\dag$ Institut de Rob{\`o}tica i Inform{\`a}tica Industrial \\
          CSIC-UPC, Barcelona, Spain, email: \mbox{porta@iri.upc.edu}\\
\\
$^\ddag$ NANO-D group at INRIA Rhône-Alpes\\
  Grenoble, France, email: \mbox{leonard.jaillet@inria.fr}
}

\date{}

\maketitle

\begin{abstract}
A well-known weakness of the probabilistic path planners is the so-called narrow
passage problem, where a region with a relatively low probability of being
sampled must be explored to find a solution path. Many strategies have been
proposed to alleviate this problem, most of them based on biasing the sampling
distribution. When kinematic constraints appear in the problem, the
configuration space typically becomes a non-parametrizable, implicit manifold.
Unfortunately, this invalidates most of the existing sampling bias approaches,
which rely on an explicit parametrization of the space to explore. In this
paper, we propose and evaluate three novel strategies to bias the sampling under
the presence of narrow passages in kinematically-constrained systems.
\end{abstract}


\section{Introduction} \label{sec:introduction}

Due to their conceptual simplicity and their efficiency in a large class of
problems, probabilistic planners provide {\em de facto} the standard solution to
the path planning problem~\cite{Lavalle_06}. Their simplicity is given by the
fact that they only require two basic operations: configuration sampling and 
connecting nearby configurations. Their efficiency largely depends on
the ability to sample in the areas that need to be crossed to connect the
start and goal configurations. Thus, the performance of these planners degrades
if the solution path needs to cross an area with a relatively low probability of being
sampled. Since these areas typically appear when the obstacles heavily restrict
the solution path, this issue is commonly known as the narrow passage problem.
The prevalent solution to address this problem is to keep the probabilistic
planning approach, but to bias the sample distribution with the aim of
increasing the number of samples in the problematic areas.

When the problem at hand includes kinematic constraints, the configuration space
is typically a non-parametrizable, implicit manifold embedded in the ambient
space defined by the variables representing the degrees of freedom of the robot.
Kinematic constraints appear quite often in practice, either due to robot's
morphology~\cite{Merlet_00}, or to geometric or contact constraints to fulfill
during operation~\cite{Rosales_IJRR11,Rodriguez_TRO08,Ballantyne_SCNA03}. In
view of their relevance, several methods for planning taking them into account
kinematic constraints have been presented
recently~\cite{Cortes_WAFR04,Stilman_TRO10,Berenson_IJRR11,Jaillet_TRO13}, but
none of them directly addresses the narrow passage problem.
The adaptation of the existing strategies to deal with this problem is not
straightforward. Most of these strategies implicitly rely on the capability of
trivially sampling new configurations and on being able to easily connect
nearby samples, typically using linear interpolation. 
Unfortunately, except for particular families of
problems~\cite{Han_WAFR00} or for problems that can be
formulated using distance constraints~\cite{Porta_ICRA2003,Porta_CK2005,Han_RSS05} 
these two basic procedures become complex
when the problem involves kinematic constraints.

This paper shows that path planners for systems with kinematic constraints can
also take advantage of sampling bias strategies to deal with the narrow passage
problem. With this objective, we propose and evaluate three novel sampling bias
strategies with increasing levels of complexity and efficiency.

After analyzing the main sampling bias methods presented in the literature in
Section~\ref{sec:related}, Section~\ref{sec:methods} describes the three
strategies introduced in this paper, Section~\ref{sec:results} evaluates them in
representative problems and, finally, Section~\ref{sec:conclusions} summarizes
the contributions and the points deserving further attention.

\section{Related Work} \label{sec:related}

The narrow passage problem has concerned researchers in motion planning
since the early works on probabilistic planners. Most of the strategies to address
this issue are based on biasing the sampling, exploiting information on the
obstacle distribution. The problem is to transfer the information from the
workspace, where obstacles are defined, to the configuration space, where the
planning is carried out.

Some methods concentrate on rigid body path
planning~\cite{Ferre_ICRA04,Zhang_ICRA08,Denny_IROS13}, where
the relation between the workspace and the configuration space is simpler. The
extension of some of these methods to articulated robots is
possible~\cite{Pan_ICRA10}, but local minima can appear in the
presence of simultaneous contacts.

Alternative approaches directly operate in the configuration space. Some of them
rely on complete representations of either the workspace~\cite{Rickert_ICRA08}
or the configuration space~\cite{Park_ICRA14}, which limits their scalability.
Other approaches are more lightweight. For instance, Gaussian
sampling~\cite{Boor_ICRA99} only requires pairs of samples where one of the
samples is in collision and the other is not. In this way, samples are only
generated near obstacles and, possibly, in narrow passages. This approach,
though, needs to directly sample valid configurations. When the problem includes
kinematic constraints, the generation of valid configurations is only simple for
particular families of problems~\cite{Han_WAFR00,Han_RSS05} or for relatively
easy problems~\cite{Zhang_ICRA13}. Thus, in general, it is complex to adapt
Gaussian sampling to the kinematically-constrained case. Approaches that repair
the solutions paths found in an environment where the free space has been
dilated also require to easily generate valid random
configurations~\cite{Shu_WAFR99}.

To detect narrow passages, the bridge-test~\cite{Sun_TRO05} focuses on
the generation of pairs of samples in collision where the central point between
these samples is collision-free. Even if the configurations could be generated
easily, this test implicitly assumes that the linear interpolation between two
samples is also a valid configuration, which is not the case when the
configuration space is an arbitrary manifold. The assumption is also taken by
the approaches that use linear interpolation to displace the random samples
toward the medial axis of the free configuration space~\cite{Lien_ICRA03}.
Moreover, approaches that use linear dimensionality reduction of the nodes
already in the RRT to bias the sampling~\cite{Dalibard_IJRR11} are also affected
by this issue.

The information of the collisions detected while extending an Rapidly-exploring
Random Tree (RRT) can be leveraged to bias the sampling~\cite{Burns_ICRA07}. For
instance, using this information the dynamic domain of an RRT node can be
defined, i.e, the part of the configuration space where it is worth to extend
the tree from that node~\cite{Yershova_ICRA05,Jaillet_IROS05}. As we will see,
this approach can be adapted to the kinematically-constrained case defining the
dynamic domain in the ambient space rather than in the configuration space. 

To the best of our knowledge, there is only one previous sampling bias method
directly designed for constrained problems~\cite{Yershova_ROMOCO09}. This work,
uses a kd-tree to generate samples, an approach that we extend herein and that
has been analyzed in detail recently for systems without kinematic
constraints~\cite{Bialkowski_IROS13}.

\section{Sampling Bias Strategies for Kinematically-Constrained Problems} \label{sec:methods}

Let consider a system described by a $n$-dimensional joint ambient
space~$\mathcal{A}$ and a $k$-dimensional configuration space $\mathcal{X}
\subset \mathcal{A}$ implicitly defined by a set of equality constraints
\begin{equation} 
  \mathcal{X}=\{{\bf x} \in \mathcal{A} \:| \:{\bf F}({\bf x})={\bf 0} \}, \label{eq:implicit}
\end{equation}
with ${\bf F}:\mathcal{A} \rightarrow \R{n-k}$,  $n>k>0$, and where we assume
that~$\mathcal{X}$ is a smooth manifold everywhere.
Let $\mathcal{O}$ be the obstacle region of~$\mathcal{X}$, such that
$\mathcal{F}=\mathcal{X}\setminus \mathcal{O} $ is the open set of the
non-colliding configurations. Let also assume that ${\bf x}_{s}$ and ${\bf
x}_{g}$ are the start and goal configurations, both in $\mathcal{F}$. Then, the
path planning problem consists of finding a collision free path linking the
query configurations while staying in~$\mathcal{F}$ i.e. to find a continuous
function $\sigma: [0,1]\rightarrow \mathcal{F}$ with $\sigma(0) = {\bf x}_{s}$,
\mbox{$\sigma(1) = {\bf x}_{g}$}. Using a probabilistic planner, a solution path
can only be efficiently determined if all the areas of~$\mathcal{F}$ to be
crossed can be effectively sampled.

Like all the probabilistic planners, general path planners for
kinematically-constrained systems  have troubles in narrow passages since they
sample either with an unspecified distribution~\cite{Berenson_IJRR11} or
approximately uniformly in~$\mathcal{X}$~\cite{Jaillet_TRO13}.
Next, we present three strategies that bias the sampling in order to alleviate
this problem. The first one is an adaptation of the dynamic domain
approach~\cite{Yershova_ICRA05}. The second one extends the kd-tree sampling
strategy proposed in~\cite{Yershova_ROMOCO09}. Finally, the third strategy
exploits the particular characteristics of the AtlasRRT planner introduced
in~\cite{Jaillet_TRO13}.

\begin{figure}[t]
\begin{center}
 \psfrag{C}[C]{$\mathcal{X}$}
 \psfrag{x1}[C]{$\mathbf{x}_s$}
 \psfrag{x2}[C]{$\mathbf{x}_n$}
 \psfrag{x3}[C]{$\mathbf{x}_r$}
 \psfrag{O}[C]{$\mathcal{O}$}
 \psfrag{R}[C]{$R$}
 \includegraphics[width=0.65\linewidth]{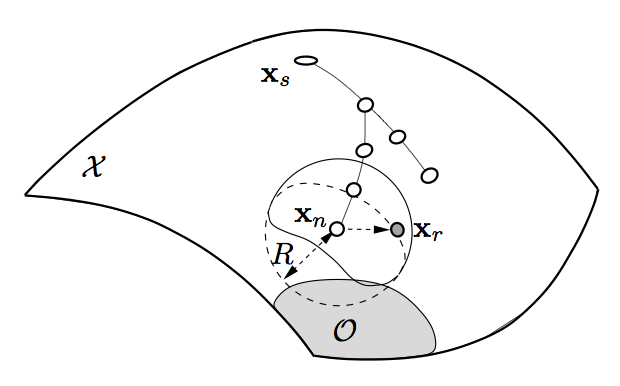}
\end{center}
\vspace{-0.5cm}
\caption{With the dynamic domain technique proposed herein, the sampling domain
         for nodes near obstacles is restricted to a ball of radius $R$ in the ambient
         space. Samples in the Voronoi region of these nodes but outside this ball are
         rejected. In the figure, the obstacle region $\mathcal{O}$ is shown in gray.
         These problematic areas are not explicitly represented, but detected with RRT
         extension failures.} \label{fig:dd}
\end{figure}

\subsection{Dynamic Domain} \label{sec:DynamicDomain}

In a standard RRT process, a node is selected for extension if a random sample
is drawn in its Voronoi region. For simplicity, the distance to determine the
Voronoi regions does not consider the obstacles. However, obstacles can severely
limit the area where the tree can actually grow from a given node.
The dynamic domain RRT planner reduces the Voronoi area of the nodes for which
an RRT extension is truncated due to a collision, i.e., of nodes close to
obstacles. The Voronoi region for those nodes is then bounded by a ball of
radius $R$. In subsequent  iterations, these nodes are extended only if a sample
is drawn in the part of the Voronoi region limited by this ball. This strategy
outperforms other RRT-like planners in cluttered scenarios because it focuses
the efforts on refining problematic areas rather than on performing unsuccessful
extensions toward obstacles. Moreover, this sampling strategy reduces the number
of collision detection tests, one of the most expensive operations for
probabilistic planners.

\begin{figure}[t]
\begin{center}
 \psfrag{C}[C]{$\mathcal{X}$}
 \psfrag{x1}[C]{$\mathbf{x}_n$}
 \psfrag{x2}[C]{$\mathbf{x}_r$}
 \psfrag{x3}[C]{$\mathbf{x}_i'$}
 \psfrag{x4}[C]{$\mathbf{x}_i$}
 \psfrag{d}[C]{$\delta$}
 \includegraphics[width=0.65\linewidth]{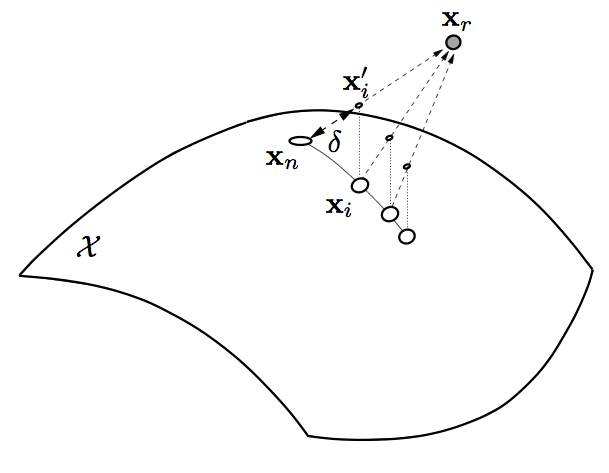}
\end{center}
\vspace{-0.5cm}
\caption{The extension procedure iteratively expands an RRT node~$\mathbf{x}_n$
         towards a random sample $\mathbf{x}_r$ in the ambient space $\mathcal{A}$ to
         obtain a new point $\mathbf{x}_i'$ that is then projected to obtain a point
         in~$\mathcal{X}$, $\mathbf{x}_i$.} \label{fig:cb-extension}
\end{figure}

Strictly speaking, when the problem includes kinematic constraints, the nearest
neighbors and, thus, the dynamic domain should be defined in~$\mathcal{X}$. This
requires to use the geodesic metric of~$\mathcal{X}$, which, for a general
manifold, is not available in closed form. In some cases, the geodesic distance
can be numerically evaluated~\cite{Memoli_JCP2001}, but the process is
computationally too expensive to be used in a practical planner. The usual
solution is to rely on the metric of~$\mathcal{A}$ as an approximation to the
metric on~$\mathcal{X}$. Herein, we propose to use this approximation to adapt
the dynamic domain approach to the kinematically-constrained case, defining the
balls associated with the RRT nodes in~$\mathcal{A}$, as illustrated in 
Fig.~\ref{fig:dd}. Since $\mathcal{X}$ is assumed to be regular manifold, a ball
around a point $\mathbf{x}$ always includes a non-null portion of $\mathcal{X}$.
This will be the effective dynamic domain for~$\mathbf{x}$. Note that, in
general, the distance in~$\mathcal{A}$ is lower than the distance in~$\mathcal{X}$
and, thus, this approximation can lead to some false positives, i.e., samples
that are accepted when they are actually too far form the corresponding RRT node
when the motion is restricted to~$\mathcal{X}$. These samples produce
unnecessary RRT extensions that can reduce the advantage of the dynamic domain
approach, but that do not compromise its completeness.

In the dynamic domain approach proposed herein, a procedure to grow RRT branches
in $\mathcal{X}$ from samples in~$\mathcal{A}$ is necessary. To this end we
propose to use the procedure introduced in~\cite{Berenson_IJRR11} and
illustrated in Fig.~\ref{fig:cb-extension}. With this procedure, a new point
$\mathbf{x}_i'$ is generated by linearly interpolating between the random sample
in~$\mathcal{A}$, $\mathbf{x}_r$, and the nearest node in the RRT,
$\mathbf{x}_n$,
\begin{equation}
 \mathbf{x}_i'=\mathbf{x}_n+ \delta \: \frac{\mathbf{x}_r-\mathbf{x}_n}{\|\mathbf{x}_r-\mathbf{x}_n\|} \:, \label{eq:interpolate}
\end{equation}
with a small parameter $\delta$. Then a configuration in $\mathcal{X}$,
$\mathbf{x}_i$, is determined initializing~$\mathbf{x}_i$  to $\mathbf{x}_i'$
and updating it with the increments
\begin{equation}
 \Delta \mathbf{x}_i = - \: \mathbf{J}(\mathbf{x}_i)\tr (\mathbf{J}(\mathbf{x}_i)\:\mathbf{J}(\mathbf{x}_i)\tr)\inv \:\mathbf{F}(\mathbf{x}_i)\:,
\end{equation}
with $\mathbf{J}(\mathbf{x}_i)$ the Jacobian of $\mathbf{F}$ evaluated at
$\mathbf{x}_i$. This correction is applied iteratively until $\mathbf{x}_i$ is
in $\mathcal{X}$, up to the numerical accuracy, or for a maximum number of
iterations. In the later case, the projection fails to converge. If successful,
the interpolation and projection steps are repeated to continue the RRT branch
until an obstacle is found or the expansion gets stalled, i.e., the distance
between two consecutive samples in the branch is too small or the projection
to~$\mathcal{X}$ fails.

Although an adaptive version of the dynamic domain planner exists where
parameter~$R$ is self-tuned~\cite{Jaillet_IROS05}, herein we will use a
fixed~$R$, which will be adjused experimentally to obtain the best performance
from this approach.

\begin{figure}[t]
\begin{center}
 \psfrag{C}[C]{$\mathcal{X}$}
 \psfrag{x1}[C]{$\mathbf{x}_s$}
 \psfrag{x2}[C]{$\mathbf{x}_n$}
 \psfrag{x3}[C]{$\mathbf{x}_r$}
 \includegraphics[width=0.65\linewidth]{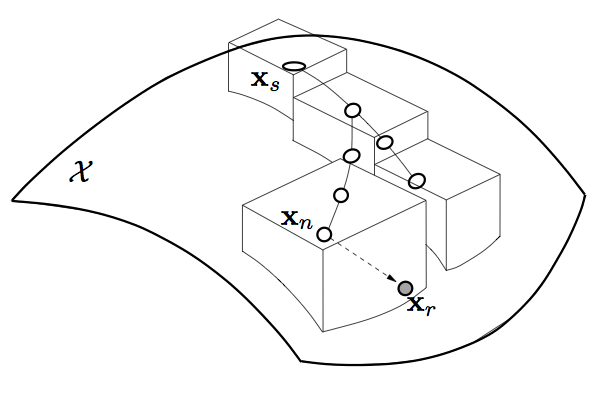}
\end{center}
\vspace{-0.5cm}
\caption{With the kd-tree sampling technique, the kd-tree used to speed up the
         search for nearest neighbors is also used for sampling. In the figure, the
         rectangles represent the \mbox{$r$-bounding} rectangles in the leaves of the
         kd-tree, which bound the area where samples are drawn.} \label{fig:kdtree}
\end{figure}

\subsection{Kd-tree Sampling} \label{sec:KDtree}

The kd-tree sampling~\cite{Yershova_ROMOCO09} aims at solving  a large class of 
constrained problems where the constraints can be formalized~as
\begin{equation}
 \|\mathbf{F}(\mathbf{x})\| \leq \varepsilon. \label{eq:kdtree}
\end{equation}
In principle \eqref{eq:kdtree} includes \eqref{eq:implicit} when
$\varepsilon=0$, but in the original approach~$\varepsilon$ is always positive
and, thus, it only deals with an approximated version of the problem addressed
herein.

In this planner the sampling domain is defined taking advantage of the kd-tree
typically used to speed up the nearest neighbor detection. A kd-tree is a
hierarchical structure that organizes a set of points in a given space
($\mathcal{A}$ in our case) by recursively partitioning this space. In the
internal nodes of the kd-tree, the space is splitted in two sub-domains along a
given dimension. If the split point is selected as the median in the split
dimension of the points to be organized, the tree is balanced. The tree leaves
include the points in the corresponding subdomain of the space. 

In this approach, samples are generated in the so-called \mbox{$r$-bounding}
rectangles. Such rectangles are the intersection between the subdomain
of~$\mathcal{A}$ covered by a given leaf and an \mbox{$n$-dimensional} box with
two opposite corners
\begin{align}
 \mathbf{x}_1 &= (\min \{ x_i^1\} - r, \ldots, \min \{ x_i^n\} - r)\tr, \\
 \mathbf{x}_2 &= (\max \{ x_i^1\} + r, \ldots, \max \{ x_i^n\} + r)\tr, 
\end{align}
with $\mathbf{x}_i = (x_i^1,\ldots,x_i^n)\tr \in L$, where $L$ is the set of
points in the leaf. The intersection with the subdomain of~$\mathcal{A}$
corresponding to the leaf ensures that the \mbox{$r$-bounding} rectangles for
different leaves do not intersect. 

The internal nodes of the kd-tree only need to add the volumes of the
\mbox{$r$-bounding} rectangles in the leaves of the corresponding sub-trees.
This information is used to distribute the samples in the sub-trees in the right
proportion. Finally, in a leaf, the sampling is uniform in the associated
\mbox{$r$-bounding} rectangle. In this way the samples are distributed uniformly
in the space covered by all the \mbox{$r$-bounding} rectangles.
Fig.~\ref{fig:kdtree} shows a mock-up of such rectangles for a set of points in
a manifold.

In the original approach, the RRT is extended toward the random sample using
linear interpolation as far as the error is below~$\varepsilon$. Thus, to avoid errors
larger than~$\varepsilon$,~$r$ must be small. Herein, we propose to combine this
sampling strategy with the extension strategy described in the previous section,
which always generates samples in~$\mathcal{X}$, i.e. with $\varepsilon=0$ up to
the numerical accuracy. In this way,~$r$ can be set to a larger value, which
favors exploration.

\begin{figure}[t]
\begin{center}
 \psfrag{C}[C]{$\mathcal{X}$}
 \psfrag{x1}[C]{$\mathbf{x}_n$}
 \psfrag{x2}[C]{$\mathbf{x}_r$}
 \psfrag{x3}[C]{$\mathbf{x}_i'$}
 \psfrag{x4}[C]{$\mathbf{x}_i$}
 \psfrag{R}[C]{$\rho_s$}
 \psfrag{d}[C]{$\delta$}

 \includegraphics[width=\linewidth]{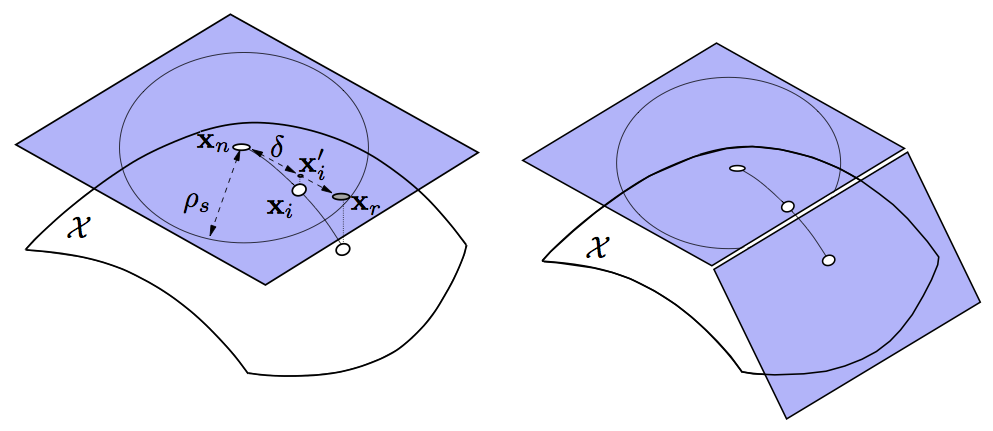} 
\end{center}
\caption{{\bf Left}: The AtlasRRT samples points in a ball defined in a tangent
         space of the manifold $\mathcal{X}$. Then, the branches are grown moving in this
         tangent space and orthogonally projecting to $\mathcal{X}$. {\bf Right}: When
         the error or the curvature of the tangent space with respect to~$\mathcal{X}$ is
         above a given threshold, or when a maximum span in a given chart is reached, a
         new tangent space is generated. This tangent space is properly coordinated with
         the previous ones, cropping the respective sampling balls.}\label{fig:continuation}
\end{figure}

\subsection{Dynamic Domain AtlasRRT} \label{sec:AtlasRRT}

The AtlasRRT~\cite{Jaillet_TRO13} does not sample in~$\mathcal{A}$, but in a
subset of the tangent bundle of~$\mathcal{X}$ that is defined as the RRT grows.
An orthonormal basis for the tangent space of~$\mathcal{X}$ at a given point
$\mathbf{x}$ is given by the $n \times k$ matrix,~${\bf \Phi}$, satisfying
\begin{equation}
 \left [ 
\begin{array}{c}
  {\bf J}({\bf x}) \\
  {\bf \Phi}\tr
\end{array}
\right] {\bf \Phi} = 
\left[
\begin{array}{c}
 {\bf 0} \\
 {\bf I}
\end{array}
\right], \label{eq:tangentSpace}
\end{equation}
with ${\bf I}$ the $k \times k$ identity matrix.
This tangent space can be seen as a chart that locally 
parametrizes~$\mathcal{X}$ and a collection of such charts forms an atlas of the
implicit manifold~\cite{Carmo_76}.

\begin{figure*}[th!]
\begin{center}
 \includegraphics[width=\linewidth]{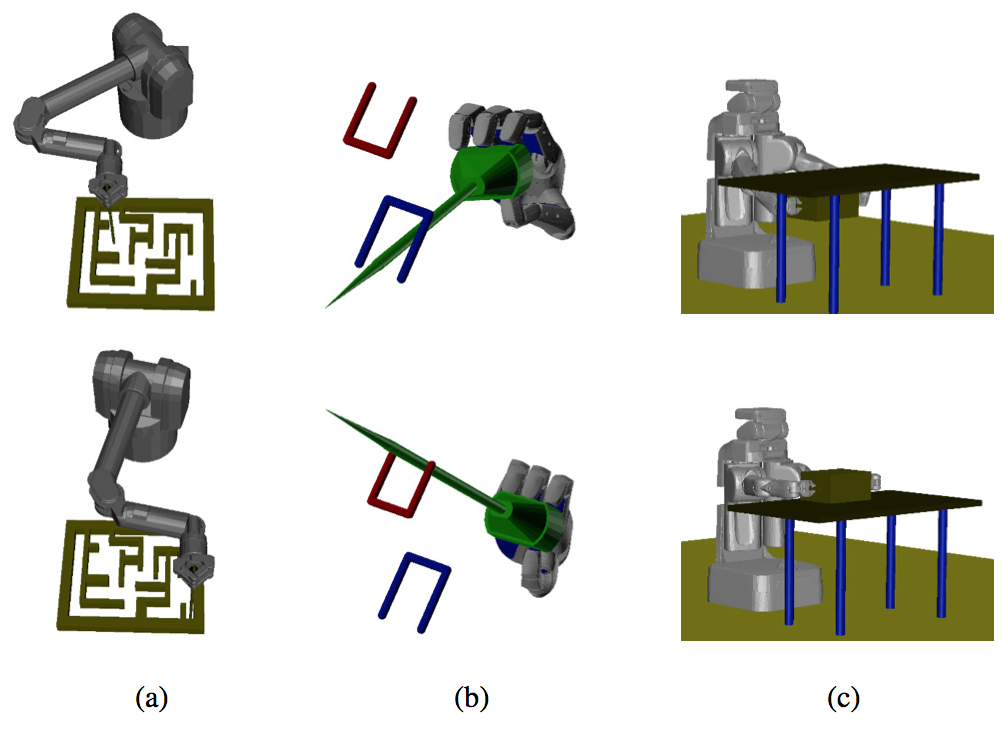} \\
\end{center}
\vspace{-0.25cm}
\caption{The three benchmarks used in this paper. For each benchmark, the top
         and bottom pictures correspond to start and goal configurations, respectively.
         (a)~The Barrett arm solving a maze problem. (b)~The Schunk anthropomorphic
         moving a needle in an environment with two U-shaped obstacles. (c)~The PR2
         service robot moving a box.} \label{fig:testcases}
\end{figure*}

To extend an AtlasRRT, a random point $\mathbf{x}_r$ is sampled in a ball of
radius $\rho_s$ in the tangent space of a given chart selected at random. Next,
a new RRT branch is grown by first linearly interpolating between $\mathbf{x}_r$
and $\mathbf{x}_n$, the node in the RRT nearest to $\mathbf{x}_r$, using
\eqref{eq:interpolate} to obtain a point $\mathbf{x}_i'$ that is then by
orthogonally projected to~$\mathcal{X}$. This projection is computed solving the
system
\begin{equation}
\left\{ \begin{array}{r}
{\bf F}({\bf x}_i) = {\bf 0}, \\ 
{\bf \Phi}_i\tr \: ({\bf x}_i-{\bf x}_i')  = {\bf 0},
 \end{array} \right.                                   \label{eq:projection}
\end{equation}
using a Newton procedure where ${\bf x}_i$ is initialized to ${\bf x}_i'$ and is
iteratively updated by the~$\Delta{\bf x}_i$ increments fulfilling
\begin{equation}
\left[
\begin{array}{c}
 {\bf J}({\bf x}_i) \\
  {\bf \Phi}_i\tr
\end{array}
\right]
 \Delta{\bf x}_i =
- \left[
\begin{array}{c}
 {\bf F}({\bf x}_i) \\
 {\bf \Phi}_i\tr \: ({\bf x}_i-{\bf x}_i')
\end{array}
\right] \;,\label{eq:newtonProjection}
\end{equation}
until the error is negligible or for a maximum number of iterations.
Fig.~\ref{fig:continuation}-left illustrates this procedure. New charts are
created when the distance or the curvature of~$\mathcal{X}$ with respect to the
tangent space being used to extend the RRT exceeds given thresholds or when
\mbox{$\|\mathbf{x}_i'-\mathbf{x}_c\|>\rho$}, with $\rho<\rho_s$ a given
parameter and $\mathbf{x}_c$ the point where the chart is tangent
to~$\mathcal{X}$. This maximum span for the charts improves the regularity of
the paving of $\mathcal{X}$. New charts are coordinated with the previous ones
to keep track of the area of~$\mathcal{X}$ already covered and the corresponding
sampling areas are cropped to avoid large overlaps between them, as shown in
Fig.~\ref{fig:continuation}-right. In subsequent sampling steps, a chart is
selected at random and random samples are drawn in the corresponding sampling
ball, but those out of the cropped area are rejected.

The natural way to adjust the sampling domain in the AtlasRRT algorithm is to
change $\rho_s$, which is a critical parameter in the original AtlasRRT planner.
Drawing inspiration from~\cite{Jaillet_IROS05}, we use a global scaling factor
that is initialized to 1 and increased by a factor \mbox{$1+\alpha$}, with
$\alpha\in(0,1)$ a fixed paremeter, when new RRT branches are grown successfully and
decreased by a factor \mbox{$1-\alpha$} when branches are interrupted due to the
presence of obstacles. In any case,~$\rho_s$ is never lower than~$\rho$, which
guarantees the continuity between the sampling areas of nearby charts and, thus,
the probabilistic completeness of the planner~\cite{Jaillet_TRO13}.

The global scaling factor can be interpreted as a balance between exploration
(when it is set to a large value) and refinement (when its value is low). This
balance is automatically adjusted according to the obstacles encountered in the
different stages of the RRT expansion. One can say that the difference
between~\cite{Jaillet_IROS05} and the dynamic domain AtlasRRT, is that the
former selects where to limit the exploration while the later decides when to
limit it since, at a given moment, the scaling factor globally affects $\rho_s$
for all charts. In practice, the global scaling of the sampling areas is more
effective than a local chart-based scaling since problematic areas are typically
covered by several charts.

\section{Experiments and Results} \label{sec:results}

The three sampling bias strategies presented in this paper have been implemented
in C and integrated in the CuikSuite~\cite{Porta_RAM14}, a general toolbox for
the motion analysis of closed-chain multi-body systems.
Figure~\ref{fig:testcases} shows the start and goal configurations for the three
benchmarks used in the experimental evaluation, which are taken
from~\cite{Jaillet_TRO13}. The first benchmark involves the Barrett
arm~\cite{BarrettArm} solving a maze problem. The stick moved by the arm has to
stay in contact with and perpendicular to the maze plane, without rotating about
its axis. In this case,~$n$, the dimension of~$\mathcal{A}$, is~9 and~$k$, the
dimension of~$\mathcal{X}$, is~3. In the second problem, the Schunk
anthropomorphic hand~\cite{Schunk2006Hand} grasps a needle which must be moved
getting around a couple of \mbox{U-shaped} obstacles, which introduce local
minima in the planning. In this benchmark,~$n$ is~23 and~$k$ is~5. In the last
benchmark, a PR2 robot with fixed base must move a box located under a table to
the top of this table, without tilting it. In this case $n$ is~16 and $k$ is~4. 
In the Barrett arm case, the maze defines a sequence of narrow passages. In the
Schunk hand example, the obstacles together with the hand self-collisions and
the strict limits of the joints also define several narrow passages. Finally, in
the PR2 benchmark there is a problematic area when the box is between the
robot and the table, but this passage is relatively wide. Thus, this example is
used to evaluate the performance of the proposed sampling strategies when the
obstacles do no particularly constraint the solution path. In the experiments
reported next, $\delta$ has been set to 0.05, a bidirectional RRT search
strategy is used, and the results are averaged over~50 runs on a Intel Core i7
at 2.93 Ghz running Mac~OS~X. Experiments running for more than 600 seconds are
considered a failure and are not taken into account to compute the average
results.

\begin{table}[t]
 \begin{center}
 \begin{tabular}{lcccccc}
                 & $R$ &     $t$ & CD tests & Col. Bran. & Succ. & Rej.    \\
 \hline
  Barrett arm    &   2 &      38 &   65000  &       0.92 &     1 & 0.96 \\
  Schunk hand    & 0.5 &     246 &  118000  &       0.78 &   0.5 & 0.99 \\
  PR2            &   2 &      18 &   15000  &       0.88 &     1 & 0.99 \\
 \hline
 \end{tabular}
 \end{center}
 \caption{Results obtained with the dynamic domain strategy.} \label{results:cbirrt-dd}
\end{table}

\begin{table}[t]
 \begin{center}
 \begin{tabular}{lcccc}
                 &     $t$ & CD tests & Col. Bran. & Succ. \\
 \hline
  Barrett arm &         47 &  110000  &       0.99 & 1     \\
  Schunk hand &        278 &  147000  &       0.83 & 0.56  \\
  PR2         &         20 &   16000  &       0.93 & 1  \\
 \hline
 \end{tabular}
 \end{center}
 \caption{Results obtained when sampling in the ambient space.} \label{results:cbirrt}
\end{table}

Table~\ref{results:cbirrt-dd} shows the results obtained with the dynamic domain
technique introduced in Secion~\ref{sec:DynamicDomain}. For each benchmark, the
table gives the value for the dynamic domain radius ($R$), the average execution
time in seconds ($t$), the number of collision detection tests (CD~test), the
ratio of RRT branches that end up in a collision (Col.~Bran.), the ratio of
successful experiments (Succ.), and, finally, the ratio of rejected samples
(Rej.). Parameter~$R$ has been set to optimize the performance of this approach.
As a reference, Table~\ref{results:cbirrt}, gives the results if the same
planning strategy is used, but sampling in~$\mathcal{A}$, without the dynamic
domain restriction.
From Tables~\ref{results:cbirrt-dd} and~\ref{results:cbirrt}, it is clear that
the dynamic domain technique reduces the number of collision
detection tests and, thus, the overall execution time. Moreover, the ratio of
RRT branches ending up in a collision is also reduced, indicating that the
dynamic domain RRT focuses more on refinement than on unsuccessful exploration.
The reduction of collision detection tests is significant in the first two
examples, where the obstacles severely constraint the solution path, and
smaller, but still a reduction, in the last problem, which is less constrained.
Note, however, that the dynamic domain technique produces a large ratio of
rejected samples, which  handicaps the approach since the generation of a valid
sample can take relatively long.

\begin{table}
 \begin{center}
 \begin{tabular}{lccccc}
                 &   $r$ &  $t$ & CD tests & Col. Bran. & Succ. \\
 \hline
  Barrett arm    &  0.25 &    18 &   50000  &       0.86 & 1     \\
  Schunk hand    &  0.5  &   220 &   97000  &       0.75 & 1     \\
  PR2            &  2    &    25 &   21000  &       0.89 & 1     \\
 \hline
 \end{tabular}
 \end{center}
 \caption{Results obtained with the kd-tree sampling strategy.} \label{results:cbirrt-kdtree}
\end{table}

Table~\ref{results:cbirrt-kdtree} shows the results obtained with the kd-tree
sampling strategy with the indicated values of $r$. This parameter is set to its
optimal value after an exhaustive search. Comparing this table with
Table~\ref{results:cbirrt-dd}, we can see that, in the first two benchmarks, the
kd-tree strategy outperforms the dynamic domain technique both in execution time
and in reduction of the number of collision detection tests. The main reasons
for this superior performance are the increased focus on refinement (as
indicated by the reduction in the ratio of branches that end up in a collision)
and the lack of rejected samples. The kd-tree sampling strategy is designed to
sample directly in the valid sampling area (i.e., in the \mbox{$r$-bounding}
rectangles at the leaves of the kd-tree) and, thus rejection is unnecessary.
This significantly speeds up this approach. However, sampling only in the
\mbox{$r$-bounding} rectangles limits the exploration, which decreases the
performance when the obstacles do not particularly constraint the solution path,
as it is the case of the PR2 benchmark. The dynamic domain strategy reaches a
better trade off between exploration and refinement since it only changes the
sampling area for the RRT nodes that are for sure close to obstacles, while 
the kd-tree sampling strategy reduces the sampling area for all nodes in 
the RRT.
Finally, note that, thanks to its focus on refinement, the kd-tree sampling
approach manages to solve the benchmarks in all the cases. This is a remarkable
result since the  dynamic domain approach is not able to solve the Schunk hand
problem in half of the runs.

\begin{table}[t]
 \begin{center}
 \begin{tabular}{lccccc}
                 &     $t$ & CD tests & Col. Bran. & Succ. & Rej. \\
 \hline
  Barrett arm    &       2 &   10000  &       0.91 &     1 & 0.84 \\
  Schunk hand    &       7 &    4800  &       0.95 &     1 & 0.64 \\
  PR2            &     1.5 &    2400  &       0.89 &     1 & 0.64 \\
 \hline
 \end{tabular}
 \end{center}
 \caption{Results obtained with the dynamic domain AtlasRRT.} \label{results:atlasrrt-adaptive}
\end{table}

\begin{table}[t]
 \begin{center}
 \begin{tabular}{lccccc}
                 &     $t$ & CD tests & Col. Bran. & Succ. & Rej. \\
 \hline
  Barrett arm    &       9 &   37000  &       0.92 &     1 & 0.97 \\
  Schunk hand    &     112 &   27000  &       0.97 &  0.48 & 0.82 \\
  PR2            &       1 &    1300  &       0.97 &     1 & 0.76 \\
 \hline
 \end{tabular}
 \end{center}
 \caption{Results obtained with the AtlasRRT.} \label{results:atlasrrt}
\end{table}

Table~\ref{results:atlasrrt-adaptive} shows the results obtained with the
dynamic domain AtlasRRT. In these experiments, $\rho_s$ is initialized to~10,
$\rho$ and $\alpha$ are set to~1 and~0.1, respectively, and the rest of
parameters are the same as those reported in~\cite{Jaillet_TRO13} for these
benchmarks. As a reference, Table~\ref{results:atlasrrt} shows the results
obtained with the same approach when~$\rho_s$ is not automatically adjusted, but
fixed to 10.
In the first two benchmarks, the dynamic domain AtlasRRT outperforms the plain
AtlasRRT in execution time, in number of collision detection tests, and in
success ratio. In the last benchmark, an aggressive exploration, i.e., fixing
$\rho_s$ to 10, produces better results. However, the performace degradation
introduced by the adaptive sampling radius strategy is minor and acceptable
taking into account that this strategy solves the problem of adjusting $\rho_s$,
which is a major issue in the original AtlasRRT. As a side-effect, the dynamic
domain AtlasRRT reduces the rejection ratio for random samples. In the AtlasRRT,
the rejected samples are those generated in a given chart, but that are out of
the sampling domain due to the coordination with neighboring charts.
When~$\rho_s$ is low, the probability of generating valid samples increases.
Finally, note that the dynamic domain AtlasRRT also outperforms the other
sampling bias strategies introduced in this paper in all aspects. This superior
performance, though,  comes at the cost of a higher conceptual and
implementation complexity.

\section{Conclusions} \label{sec:conclusions}

This paper introduces and evaluates three strategies to deal with the narrow
passage problem in systems with kinematic constraints, a relevant problem hardly
addressed in the literature so far. The three proposed strategies are based on
biasing the sampling distribution and produce improvements with respect to the
equivalent planners with unbiased sampling, increasing the ratio of successful
runs and reducing the number of collision detection tests and, thus, reducing
the execution time. This is achieved by increasing the focus on the refinement
of the RRT tree, rather than on its fast expansion, which, in cluttered spaces,
mainly leads to collisions. The proposed systems do not rule out exploration,
but they reduce it where or when necessary, depending on the approach. 

The presented results show that the dynamic domain AtlasRRT strategy is the one
that provides better performance, but it is significantly more complex than the
two other strategies presented herein. However, we provide an open source
implementation of the three approaches with the aim of facilitating their
integration in new planners~\cite{CuikWeb}.

When the problem involves kinematic constraints, the distance between
configurations should be based on the intrinsic metric of the implicit manifold.
However, for a general manifold, this metric is complex to compute. Approximated
nearest neighbor methods for points on manifolds have been recently
introduced~\cite{Chaudhry_ECCV2010} and we plan to integrate them in our planner
in the near future. Moreover, in this paper, we only considered system with
kinematic constraints. In the future, we would like to deal with dynamic
constraints too. The presence of these constraints also define an implicit
manifold that need to be efficiently explored and, in this exploration, the
sampling bias also plays a fundamental role~\cite{Shkolnik_ICRA09}.

\bibliographystyle{plain}

\end{document}